\documentclass{article}

\usepackage{arxiv}

\usepackage[utf8]{inputenc} 
\usepackage[T1]{fontenc}    
\usepackage{hyperref}       
\usepackage{url}            
\usepackage{booktabs}       
\usepackage{amsfonts}       
\usepackage{nicefrac}       
\usepackage{microtype}      
\usepackage{lipsum}
\usepackage{xcolor}
\usepackage{listings}
\usepackage{bm}
\usepackage{graphicx}
\usepackage[linesnumbered,vlined]{algorithm2e}

\definecolor{codegreen}{rgb}{0,0.5,0}
\definecolor{codepurple}{rgb}{0.88,0,0.32}
\definecolor{backcolour}{rgb}{0.97,0.97,0.97}
 
\lstdefinestyle{coding}{
    backgroundcolor=\color{backcolour},   
    commentstyle=\color{codegreen},
    keywordstyle=\color{codepurple},
    basicstyle=\ttfamily,
    breakatwhitespace=false,         
    breaklines=true,                 
    captionpos=b,                    
    keepspaces=true,                 
    numbers=none,                    
    numbersep=5pt,                  
    showspaces=false,                
    showstringspaces=false,
    showtabs=false,                  
    tabsize=4
}
 
\title{Opytimizer: A Nature-Inspired Python Optimizer}

\author{
  Gustavo H. de Rosa, Douglas Rodrigues, João P. Papa \\
  Department of Computing \\
  São Paulo State University \\
  Bauru, São Paulo - Brazil \\
  \texttt{gustavo.rosa@unesp.br, douglasrodrigues.dr@gmail.com, joao.papa@unesp.br} \\
}
\begin{document}

\maketitle

\begin{abstract}
Optimization aims at selecting a feasible set of parameters in an attempt to solve a particular problem, being applied in a wide range of applications, such as operations research, machine learning fine-tuning, and control engineering, among others. Nevertheless, traditional iterative optimization methods use the evaluation of gradients and Hessians to find their solutions, not being practical due to their computational burden and when working with non-convex functions. Recent biological-inspired methods, known as meta-heuristics, have arisen in an attempt to fulfill these problems. Even though they do not guarantee to find optimal solutions, they usually find a suitable solution. In this paper, we proposed a Python-based meta-heuristic optimization framework denoted as Opytimizer. Several methods and classes are implemented to provide a user-friendly workspace among diverse meta-heuristics, ranging from evolutionary- to swarm-based techniques.
\end{abstract}

\keywords{Python \and Optimization \and Meta-Heuristics}

\section{Introduction}
\label{s.intro}

Mathematics is a formal basis for the representation and solution to the most diverse tasks, being applied throughout several fields of interest. Economy, Geophysics, Electrical Engineering, Mechanical Engineering, Civil Engineering, among others, are some examples of areas that use mathematical concepts in order to solve their challenges. The necessity of finding a suitable set of information (parameters) to solve a specific application fostered the study of mathematical programming, commonly known as optimization. A classic optimization task is the Traveling Salesman problem~\cite{papadimitriou77}, which consists of visiting $m$ locations with a minimum cost path\footnote{Each location can only be visited once, and then it must return to the origin point.}. Additionally, it is possible to create analogous situations with daily face-offs, i.e., a logistics company needs to find minimum-cost paths to transport their cargos, thus minimizing their expenses. Furthermore, there are other diverse optimization problems that one can face, such as industrial components modeling~\cite{rao09}, operations research~\cite{rardin98}, market economic models~\cite{konno91}, molecular modeling~\cite{barone98}, among others.

\begin{figure}[!ht]
\centering
\includegraphics[scale=0.28]{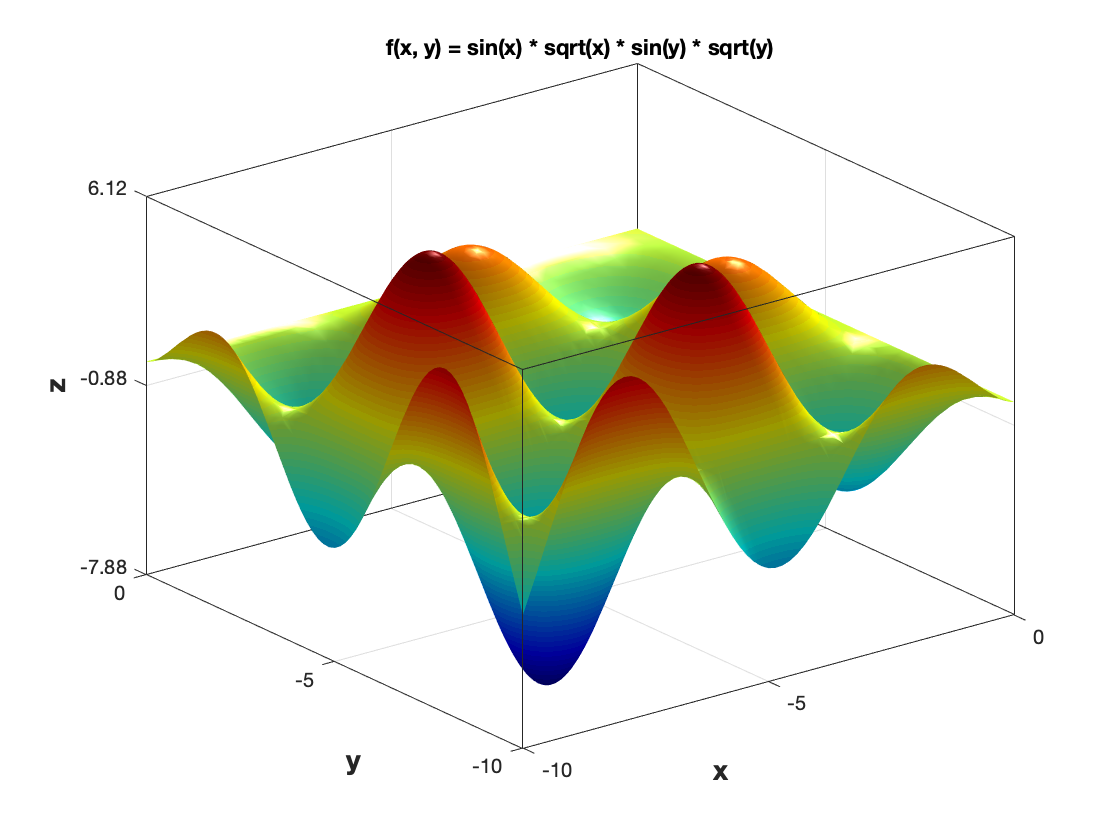}
\caption{Example of a mathematical function with several local optimal points.}
\label{f.math_function}    
\end{figure}

An optimization problem consists in maximizing or minimizing a function through a systematic choice of possible values to the problem. In other words, the optimization procedure finds the most suitable values of a fitness function, given a pre-defined domain. Equation~\ref{e.min_opt} describes the minimum\footnote{Remember that $\max(f) = \min(-f)$.} generic optimization model without constraints.

\begin{equation}
\label{e.min_opt}
\min_{\forall \bm{x} \in {\cal{R}}^n} f(\bm{x})
\end{equation}

where $f(\bm{x})$ stands for the fitness function, while $\bm{x} \in {\cal{R}}^n$. The optimization of this function aims at finding the most suitable set of values for $\bm{x}$, denoted as $\bm{x}^*$, where $f(\bm{x}^*) \leq f(\bm{x})$. Nevertheless, when dealing with more complex fitness functions, several minima or maxima points (local optima) appears and difficult the search for the optimal points. Figure~\ref{f.math_function} illustrates an example of this situation.

Traditional optimization methods~\cite{bertsekas99}, such as the iterative methods, e.g., Newton method, Quasi-Newton method, Gradient Descent, and Interpolation methods, use the evaluation of gradients and Hessians, being unfeasible to several applications due to their computational burden. In the past decades, an exciting proposition denoted as meta-heuristic has been employed to solve several optimization problems. A meta-heuristic technique consists of a high-level procedure projected to generate or select a heuristic, which provides a sufficiently feasible solution to the optimization problem. Moreover, a meta-heuristic is a procedure that combines the concepts of \emph{exploration}, used to perform searches throughout the search space, and \emph{exploitation}, which is used to refine a promising solution based on its neighborhood. Essentially, meta-heuristic techniques~\cite{yang_review} are strategies that guide the process of searching for quasi-optimum solutions. They are mostly constituted of simple local searches and complex learning procedures, usually inspired by biological behaviors. Additionally, they are non-domain specific and have mechanisms to avoid being trapped in local optima points.

In this paper, we propose a nature-inspired open-source Python optimization library focused on minimization problems, called Opytimizer\footnote{https://github.com/gugarosa/opytimizer}. Mainly, the idea is to provide a user-friendly environment to work with function optimization and meta-heuristic techniques by creating high-level methods and classes, removing from the user the burden of programming at a mathematical level.  The main contributions of this paper are threefold: (i) to introduce a meta-heuristic optimization library in the Python language, (ii) to provide an easy-to-go implementation, and several integrations with other frameworks, and (iii) to fill the lack of research regarding meta-heuristic optimizations.

The remainder of this paper is organized as follows. Section~\ref{s.literature} provides a literature review and related works concerning meta-heuristic optimization frameworks. Section~\ref{s.theory} introduces a theoretical background concerning agents, search spaces and meta-heuristics. Section~\ref{s.opytimizer} presents concepts of the Opytimizer library, such as its architecture, and an overview of the included packages. Section~\ref{s.library} provides more deeply concepts about the library, such as how to install, how to read its documentation, some pre-included examples, and how to run unitary tests. Furthermore, Section~\ref{s.applications} provides basic information about the usage of the library, i.e., how to run pre-defined examples, how to integrate with other frameworks, and how to model a new experiment. Finally, Section~\ref{s.conclusion} states conclusions and future works.

\section{Literature Review and Related Works}
\label{s.literature}

Meta-heuristics algorithms have arisen as a new approach to tackle optimization problems. They offer simple implementations along with complex search mechanisms that guarantee to find a solution and avoid being trapped in local optimums. It is possible to find their usage in a wide range of applications, such as machine learning fine-tuning~\cite{RosaCIARP:15,PapaJoCS:15}, engineering~\cite{Yang:10, OhASC:17}, medical science~\cite{Klein:07,Dey:14}. For instance, Yang et al.~\cite{YangCS:10} proposed a new algorithm called Cuckoo Search in an attempt to solve engineering design optimization problems, achieving better results than the traditional Particle Swarm Optimization. Roos et al.~\cite{Roos:10} used concepts of evolutionary populations to interplay between risk-taking and sequentially of choices, creating more robust state-dependent models. Furthermore, Papa et al.~\cite{PapaGECCO:15} proposed a hyperparameter selection of Restricted Boltzmann Machines through a Harmony Search algorithm, achieving better results than empirically setting the hyperparameters. Moreover, Papa et al.~\cite{PapaASC:16} extended the previous approach in the context of Deep Belief Networks. Recently, Reyes et al.~\cite{Reyes:18} proposed a multi-objective optimization strategy to perform batch-mode active learning based on an evolutionary algorithm, which turned out to be a more effective approach.

Even though numerous works in the literature use meta-heuristic techniques as an optimization algorithm, the literature still lacks works regarding meta-heuristic frameworks or open-sourced libraries. There are several implementations among the internet, such as EvoloPy\footnote{https://github.com/7ossam81/EvoloPy} and MetaheuristicAlgorithmsPython\footnote{https://github.com/tadatoshi/metaheuristic\_algorithms\_python}, but they only provide straightforward implementations of the algorithms, not packing them into libraries or packages, nor providing the necessary tools for users to design new experiments or integrate with other frameworks. Additionally, they lack documentation and test suites, which helps users in understanding the code and implementing new methods and classes.

On the other hand, Inspyred\footnote{https://github.com/aarongarrett/inspyred} tries to provide a more robust library-look, even though it lacks documentation, test suites, maintenance, and several state-of-the-art meta-heuristic algorithms. Moreover, an outstanding library is the PonyGE2~\cite{Fenton:17},  which deals purely with Grammatical Evolution and provides a fast-prototyping environment for students and researches. Furthermore, one of the best packages that one can find is DEAP~\cite{DEAP:12}, which is currently a state-of-the-art package concerning Evolutionary Computation. Notwithstanding, one might find it challenging to work with and integrate with other frameworks, such as Tensorflow or PyTorch. Additionally, apart from evolutionary-based algorithms, DEAP lacks implementing hypercomplex representations, some swarm-based algorithms, and even the newest trending meta-heuristics, such as Bat Algorithm, Cuckoo Search, and Firefly Algorithm.

Another state-of-the-art library concerning meta-heuristic optimization techniques is the LibOPT~\cite{Papa:17}. Although it provides a wide range of algorithms and benchmarking functions, the library is implemented in C language, making it extremely difficult to integrate with other frameworks or packages, primarily because the machine learning community is turning their attention to the Python language. Additionally, there is no documentation nor test suites to help users throughout their usage experience.

Therefore, Opytimizer attempts to fill the gaps concerning meta-heuristic optimization frameworks. It is purely implemented in Python, only using Numpy and Matplotlib as its dependencies. It provides extensive documentation, test suites, several pre-loaded examples, and even integrations with other well-known frameworks. Moreover, every line of code is commented, there are continuous integration tests for every new push to its repository, an amazing readme which explains on how to get started with the library and full-time maintenance and support.

\section{Theoretical Foundation}
\label{s.theory}

Before diving into Opytimizer's library, we present a theoretical foundation to assist one in understanding how its basic structures work. In the next subsections, we mathematically explain how agents, search spaces, and meta-heuristics works, as well as, we provide pseudo-algorithms to contextualize their mechanisms.

\subsection{Agents}
\label{ss.math_agents}

An agent is the first entity in the meta-heuristic optimization workflow, providing valuable knowledge about the optimization problem itself. In other words, an agent is an individual (particle) responsible for traversing a search space in an attempt to find a feasible solution to the optimization task. Additionally, one can find several names for agents in distinct meta-heuristic techniques, such as bats for the Bat Algorithm, nests for the Cuckoo Search, fireflies for the Firefly Algorithm, and particles for the Particle Swarm Optimization, among others.

Mathematically speaking, let $a$ be an agent, composed of two properties: $\bm{x}$, where $\bm{x} \in \Re^{n \times d}$ is a real-valued position tensor of size equal to number of decision variables $\times$ number of dimensions (e.g., complex, quaternion, octonion), and $f$, which is the agent's fitness value and initially instantiated as the maximum real value possible.

\subsection{Search Spaces}
\label{ss.math_ss}

A search space composes the search structure itself, holding all possible values that can be applied to the fitness function. Essentially, a search space is formed by several agents, which are responsible for traversing and finding a feasible solution to a particular problem. Moreover, the search space encodes the lower and upper bounds for each decision variable depicted in the problem, obligating the agents to search on a fixed window. Furthermore, the search space encodes the number of search iterations that will be performed, as well as a pointer to the current best agent (lowest fitness value in minimization tasks).

Let $s$ be a search space, composed of five properties: $\bm{A}$, which represents a vector of $m$ agents that composes the space, $a^{*}$, which is the pointer to the best agent, $T$, which is the number of search iterations, $\bm{b}_l$ and $\bm{b}_u$, where $\bm{b}_l$ and $\bm{b}_u \in \Re^n$ represents the lower and upper bounds for each decision variable, respectively.

\subsection{Meta-Heuristics}
\label{ss.heuristics}

As aforementioned, meta-heuristics are high-level search strategies that combine the power of exploration and exploitation. In other words, the meta-heuristic is the searching algorithm itself, mostly composed of simple local searches and complex learning procedures that attempts to find the most suitable solution for a particular problem. Each meta-heuristic has its searching strategy and is usually inspired by a biological phenomenon, such as the bats' behavior in the Bat Algorithm, the fireflies' behavior in the Firefly Algorithm, the water cycle in the Water Cycle Algorithm, and the flowers' pollination process in the Flower Pollination Algorithm, among others. Algorithm~\ref{a.meta_heuristic} depicts a pseudo-algorithm of how a meta-heuristic works\footnote{Note that this is only a pseudo-algorithm and may not correspond to a real meta-heuristic procedure.}.

\begin{algorithm}[!h]
\KwIn{Vector of agents $A$, number of maximum iterations $T$, lower and upper bounds $\bm{b}_l$ and $\bm{b}_u$.}
\KwOut{A solution tensor $\bm{x}^{*}$ holding the most feasible positions for the task's fitness function.}
Define a fitness function $f(\bm{x})$\;
Initialize all agents $a_i \in A$, where $i = \{1, 2, \dots, m\}$\;
$t = 1$\;
\While{$(t <= T)$}{
	\For{$i$ from $1$ to $m$}{
		Perform searching strategy and update $a_i \rightarrow \bm{x}$\;
		Check if $a_i \rightarrow \bm{x}$ is between $\bm{b}_l$ and $\bm{b}_u$\;
		\If{$(f(a_i \rightarrow \bm{x}) < f(a^{*} \rightarrow \bm{x}))$}{
			$a^{*} = a_i$\;
		}
	}
	$t = t+1$\;
}
$\bm{x}^{*} = a^{*} \rightarrow \bm{x}$\;
\caption{Meta-heuristic pseudo-algorithm.}
\label{a.meta_heuristic} 
\end{algorithm}

\section{Opytimizer}
\label{s.opytimizer}

\begin{figure}[!ht]
\centering
\includegraphics[scale=0.4]{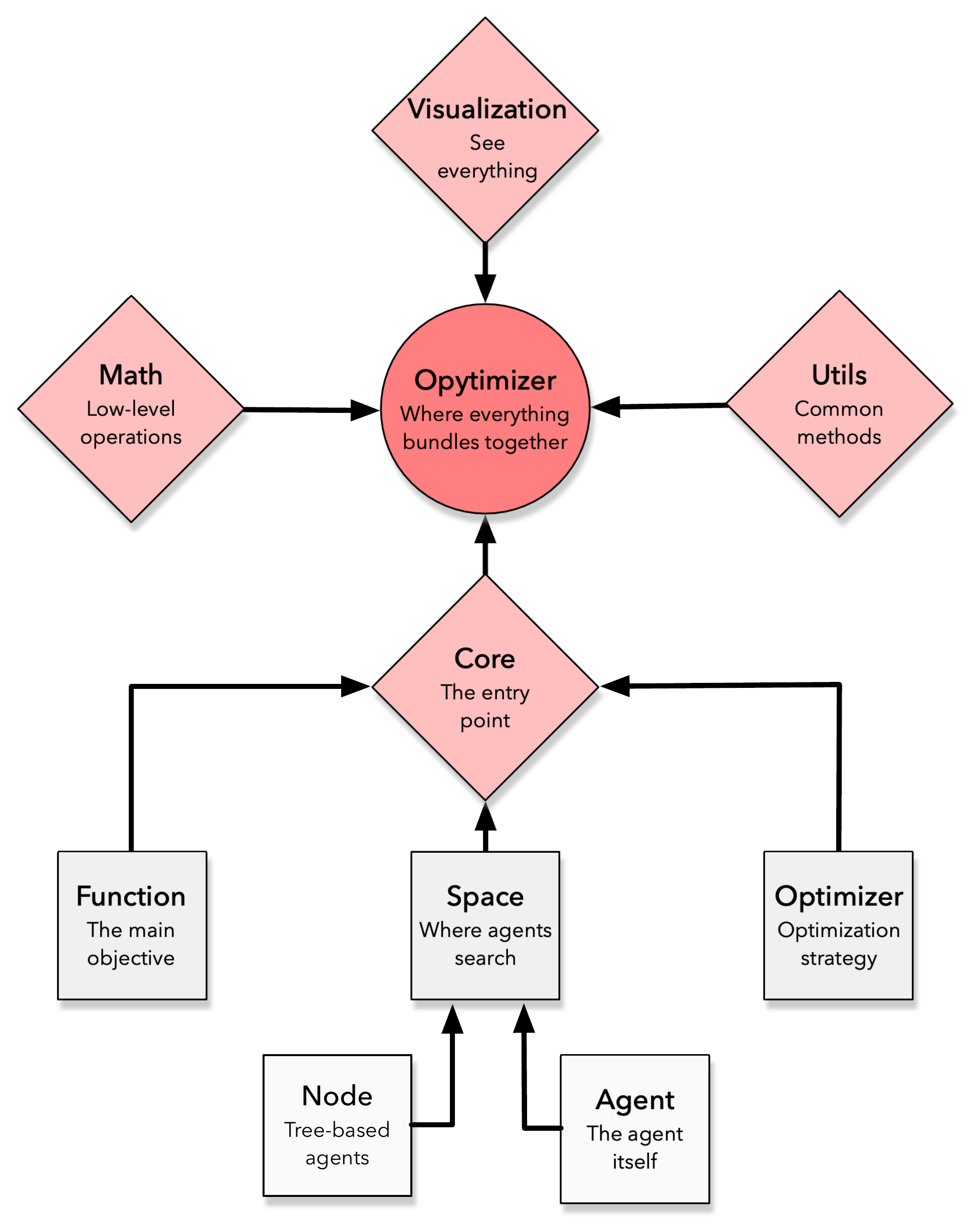}
\caption{Flowchart of Opytimizer's architecture.}
\label{f.flowchart}    
\end{figure}

Opytimizer is divided into several packages, each one being responsible for particular classes and methods across the library. Figure~\ref{f.flowchart} illustrates an overview of Opytimizer's architecture, while the next sections provide within more details each one of its packages.

\subsection{Core}
\label{ss.core}

\begin{figure}[!ht]
\centering
\includegraphics[scale=0.5]{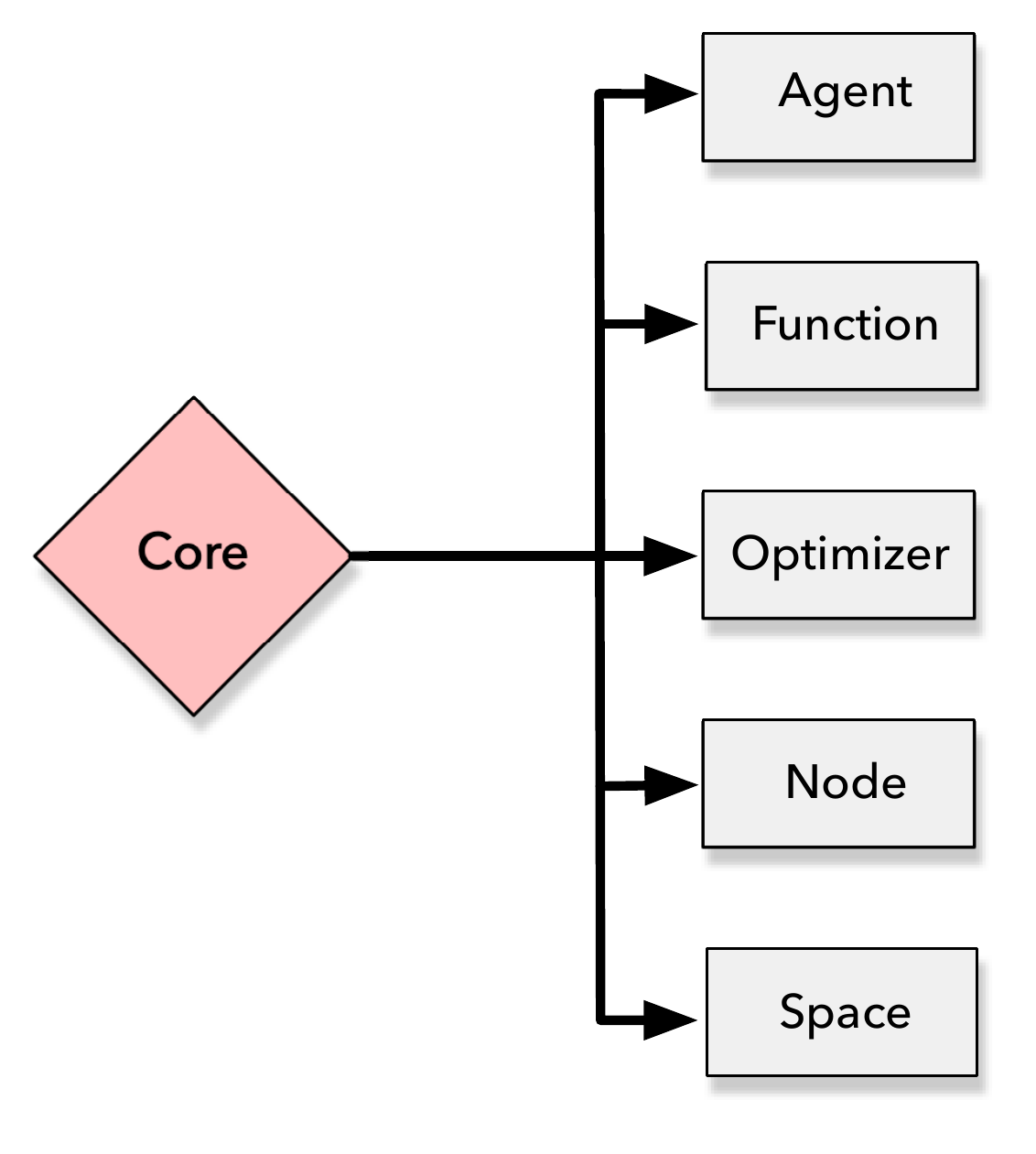}
\caption{Flowchart of Opytimizer's core package.}
\label{f.flowchart_core}    
\end{figure}

The core package works as the parent of all sub-classes that Opytimizer provides. It serves as a building block for defining more specific structures that one may need when designing a meta-heuristic algorithm or a new search strategy. As depicted in Figure~\ref{f.flowchart_core}, five modules bundles into the core package, as follows:

\begin{itemize}
\item \textbf{Agent:} An agent is a basic class that represents the individual of an optimization task, constituted of a fitness value and its position on the search space. It is a generic nomenclature as each algorithm might have an individual agent, such as bats, fireflies, particles, among others;

\item \textbf{Function:} The function is a wrapper to the objective function that will be evaluated throughout the optimization task, e.g., benchmarking function, a classifier's accuracy over a validation set, or the reconstruction error of a machine learning algorithm;

\item \textbf{Node:} When working with tree-based evolutionary algorithms, such as Genetic Programming, we need an additional class to represent the tree itself. The Node class arises to tackle this issue by providing an easy-implementation of a tree structure;

\item \textbf{Optimizer:} In Opytimizer, every meta-heuristic technique inherits from the Optimizer class. Essentially, it is a wrapper that provides shared properties and methods among all meta-heuristic algorithms;

\item \textbf{Space:} Agents or Nodes needs to be bundled into a search space. Nevertheless, as each training strategy might need a different search space, we offer a building block to implement different possibilities. The Space class provides properties and methods that encode and initialize agents or nodes throughout the space. 
\end{itemize}

\subsection{Functions}
\label{ss.functions}

\begin{figure}[!ht]
\centering
\includegraphics[scale=0.5]{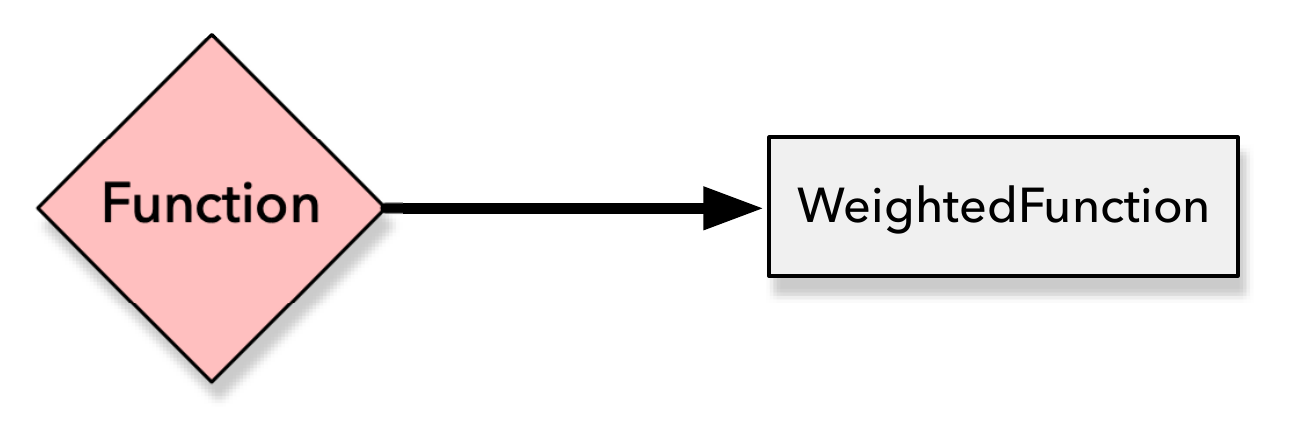}
\caption{Flowchart of Opytimizer's function package.}
\label{f.flowchart_function}    
\end{figure}

When designing an optimization task, one might need more complex implementations than single-objective functions. The functions package provides classes and methods to compose high-level abstract functions strategies, such as multi-objective and many-objective ones. Currently, Opytimizer offers only an additional multi-objective weight-based strategy, which is illustrated by Figure~\ref{f.flowchart_function} and described as follows:

\begin{itemize}
\item \textbf{WeightedFunction:} When designing a multi-objective task, one can use the WeightedFunction class and plug several objective functions and weigh them into a single objective.
\end{itemize}

\subsection{Math}
\label{ss.math}

\begin{figure}[!ht]
\centering
\includegraphics[scale=0.5]{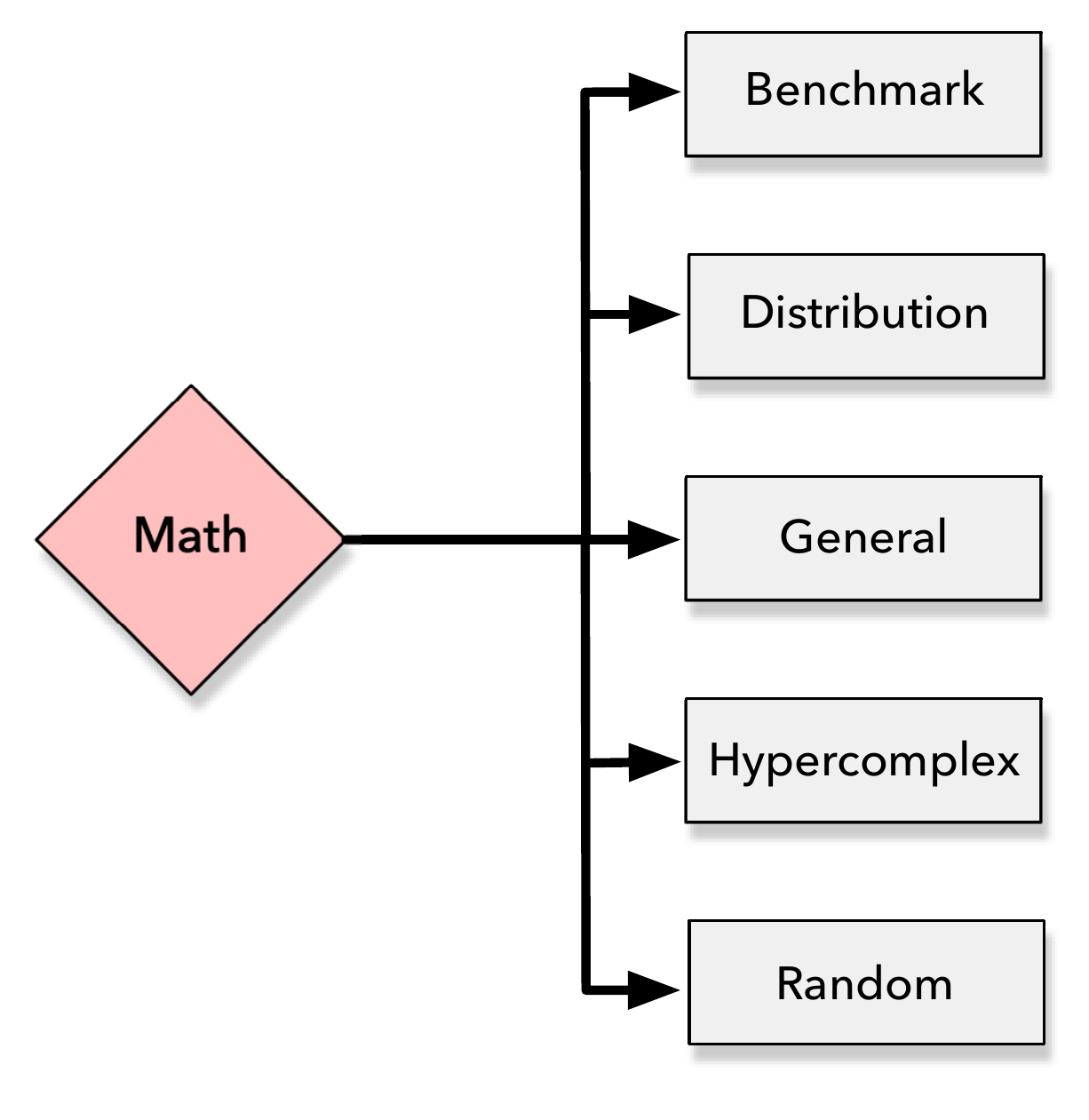}
\caption{Flowchart of Opytimizer's math package.}
\label{f.flowchart_math}    
\end{figure}

In order to ease the user's life, Opytimizer offers a mathematical package, containing low-level math implementations, illustrated by Figure~~\ref{f.flowchart_math}. Essentially, some repetitive methods that are used throughout the library are defined in this package, as follows:

\begin{itemize}
\item \textbf{Benchmark:} Benchmarking functions are mathematical functions that help to validate optimization algorithms. This module is composed of several functions with particular traits, such as modality, basins, valleys, separability, and dimensionality;

\item \textbf{Distribution:} Some algorithms might need particular distributions throughout their implementation. This module tackles this issue by providing methods to generate distinct mathematical distributions, such as Bernoulli's or L\'evy's;

\item \textbf{General:} Common-use methods that do not have a particular category are defined in this module;

\item \textbf{Hypercomplex:} When dealing with hypercomplex numbers, such as quaternions or octonions, one might need some additional methods to interpret them into a real-valued space;

\item \textbf{Random:} Lastly, several algorithms use random numbers for sampling or defining some heuristics. This module is capable of generating uniform and Gaussian random numbers.
\end{itemize}

\subsection{Optimizers}
\label{ss.optimizers}

\begin{figure}[!ht]
\centering
\includegraphics[scale=0.5]{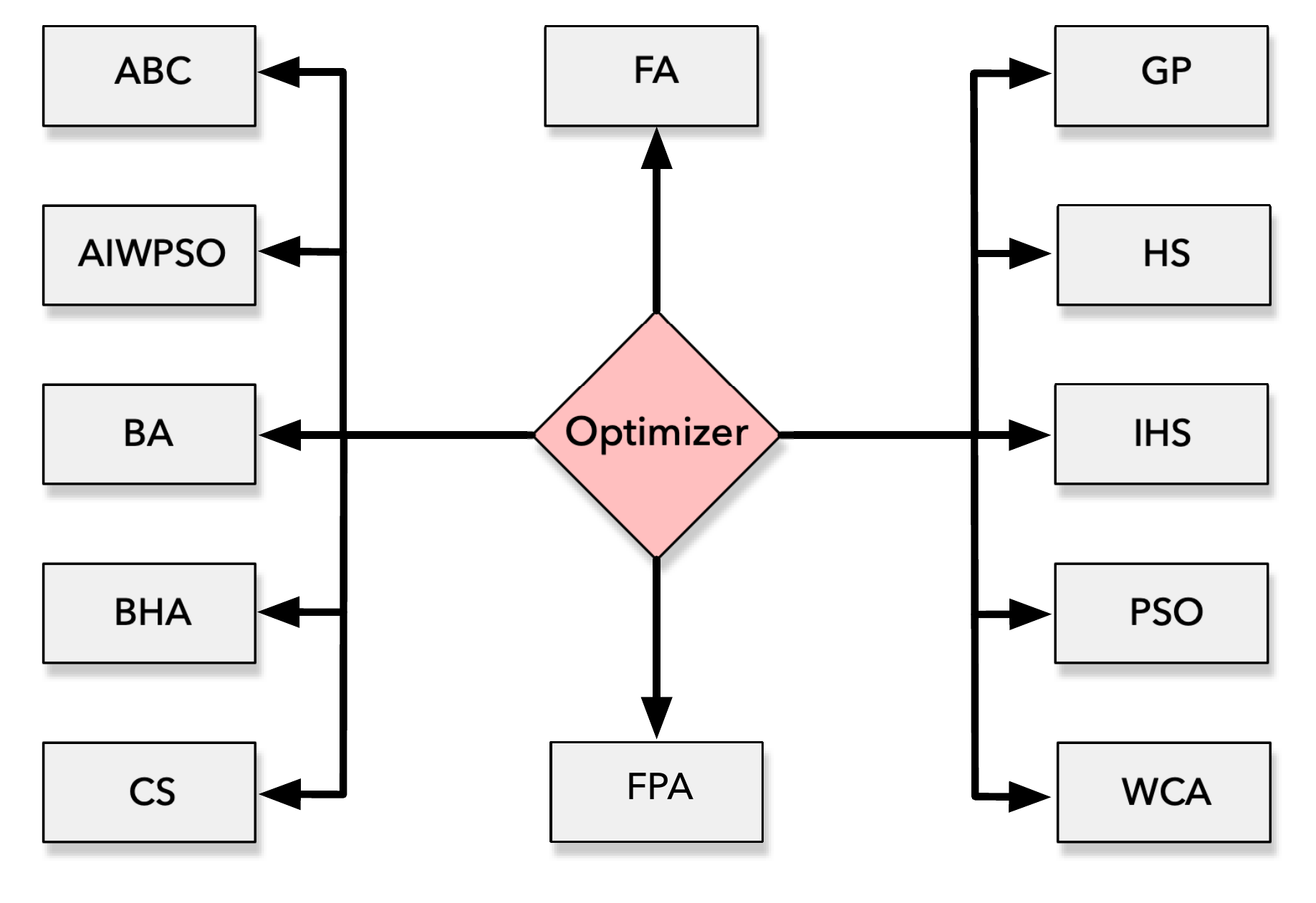}
\caption{Flowchart of Opytimizer's optimizer package.}
\label{f.flowchart_optimizer}    
\end{figure}

The Optimizer package is the heart of the heuristics, where it is possible to find a large number of meta-heuristics implementations, which are depicted by Figure~\ref{f.flowchart_optimizer}. Additionally, we provide their full names, acronyms, and references as follows:

\begin{itemize}
	\item \textbf{ABC:} Artificial Bee Colony~\cite{Karaboga:07};
	\item \textbf{AIWPSO:} Adaptive Inertia Weight Particle Swarm Optimization~\cite{Nickabadi:11};
	\item \textbf{BA:} Bat Algorithm~\cite{YangBA:12};
	\item \textbf{BHA:} Black Hole Algorithm~\cite{Hatamlou:13};
	\item \textbf{CS:} Cuckoo Search~\cite{YangCS:10};
	\item \textbf{FA:} Firefly Algorithm~\cite{YangFFA:10};
	\item \textbf{FPA:} Flower Pollination Algorithm~\cite{YangFPA:14};
	\item \textbf{GP:} Genetic Programming~\cite{Koza:92};
	\item \textbf{HS:} Harmony Search~\cite{Geem:01};
	\item \textbf{IHS:} Improved Harmony Search~\cite{Mahdavi:07};
	\item \textbf{PSO:} Particle Swarm Optimization~\cite{Kennedy:01};
	\item \textbf{WCA:} Water Cycle Algorithm~\cite{Eskandar:12}.
\end{itemize}

\subsection{Space}
\label{ss.space}

\begin{figure}[!ht]
\centering
\includegraphics[scale=0.5]{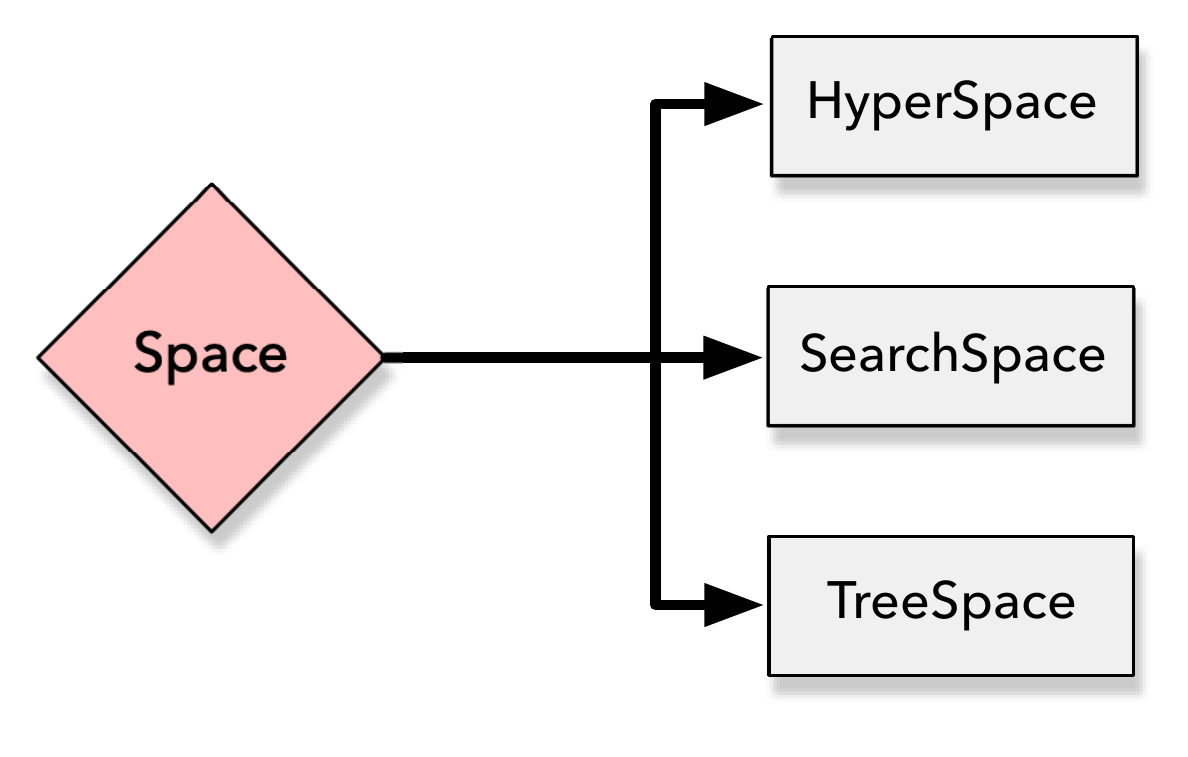}
\caption{Flowchart of Opytimizer's space package.}
\label{f.flowchart_space}    
\end{figure}

One can see the space as the place that agents update their positions and evaluate a fitness function. Furthermore,  each strategy may consider a different type of space, exhibited by Figure~\ref{f.flowchart_space}. Thus, we are glad to support diverse space implementations, such as follows:

\begin{itemize}

\item \textbf{HyperSpace:} When using a hypercomplex representation, it is important to define some constraints and specificities that are related to this kind of strategy, such as mapping agents' components between 0 and 1 and further spanning them into real-valued numbers;

\item \textbf{SearchSpace:} The traditional search space representation, where agents are mapped between lower and upper bounds and perform their search in a real-valued space;

\item \textbf{TreeSpace:} When using tree-based evolutionary algorithms, one needs to employ the TreeSpace as it holds properties and methods related to producing trees, as well as searching and evaluating them. 

\end{itemize}

\subsection{Utils}
\label{ss.utils}

\begin{figure}[!ht]
\centering
\includegraphics[scale=0.5]{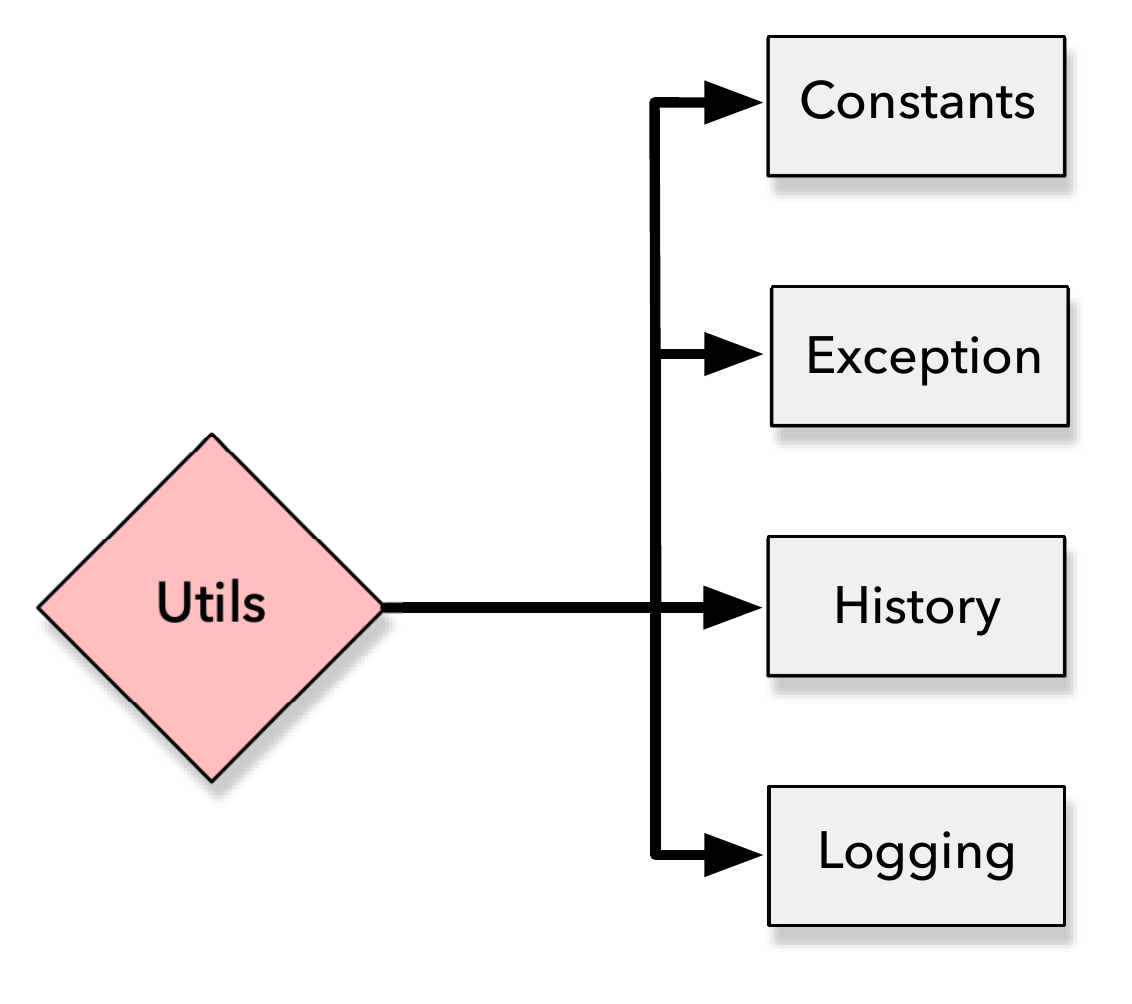}
\caption{Flowchart of Opytimizer's utils package.}
\label{f.flowchart_utils}    
\end{figure}

A utility package provides common tools that are shared across the application, as it is better only to implement once and re-use it across other modules as shown in Figure~~\ref{f.flowchart_utils}. This package provides the following modules:

\begin{itemize}

\item \textbf{Constants:} Constants are fixed numbers that do not change throughout the implementation. For the sake of easiness, they are implemented in the same module;

\item \textbf{Exception:} In order to help users when designing their experiments, the exception module provides common errors and exceptions that might occur when invalid settings are used in Opytimizer;

\item \textbf{History:} After running an optimization task, one might need some additional information about the agents or the search space. The History module implements a history-based object that saves everything that one might need, such as agents' positioning and their fitness;

\item \textbf{Logging:} Every method or action that is invoked in the library are logged into a log file. One can follow the log in order to find potential mistakes or even important warnings or success messages throughout the optimization task.

\end{itemize}

\subsection{Visualization}
\label{ss.visualization}

\begin{figure}[!ht]
\centering
\includegraphics[scale=0.5]{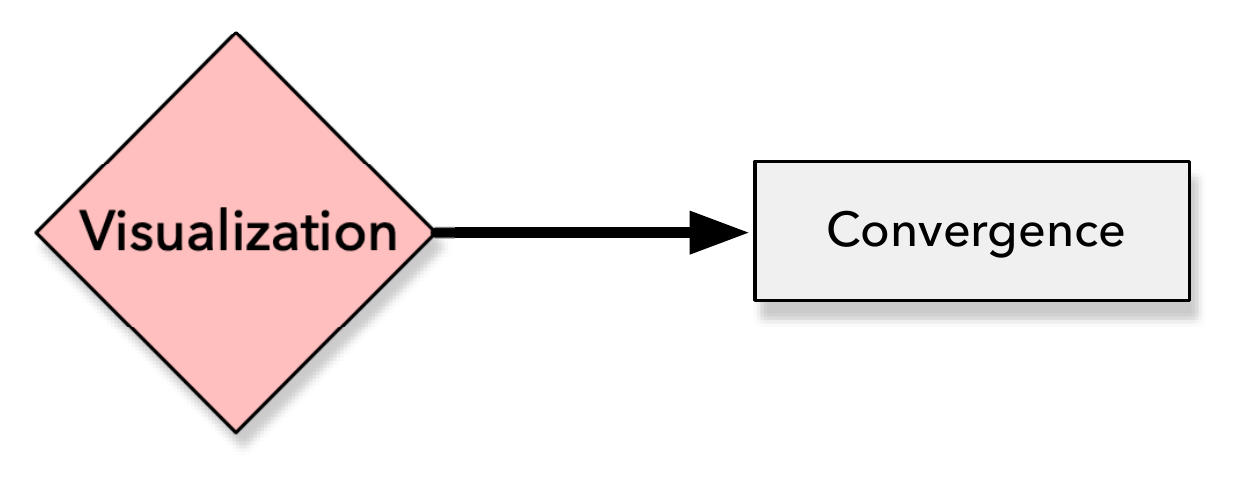}
\caption{Flowchart of Opytimizer's visualization package.}
\label{f.flowchart_visualization}    
\end{figure}

\begin{figure}[!ht]
\centering
\includegraphics[scale=0.65]{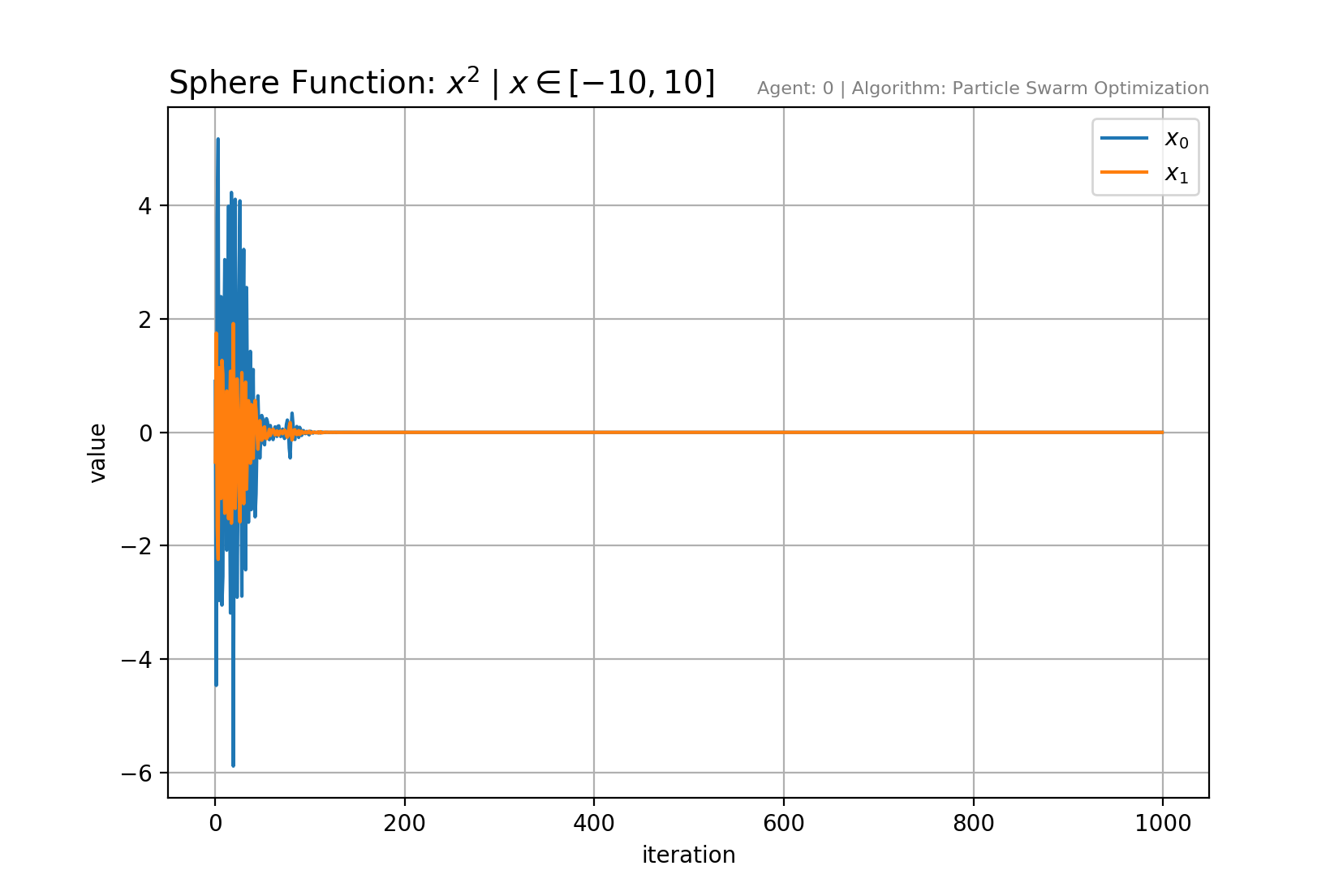}
\caption{Positions' convergence plot using the visualization package.}
\label{f.convergence_pos}    
\end{figure}

\begin{figure}[!ht]
\centering
\includegraphics[scale=0.65]{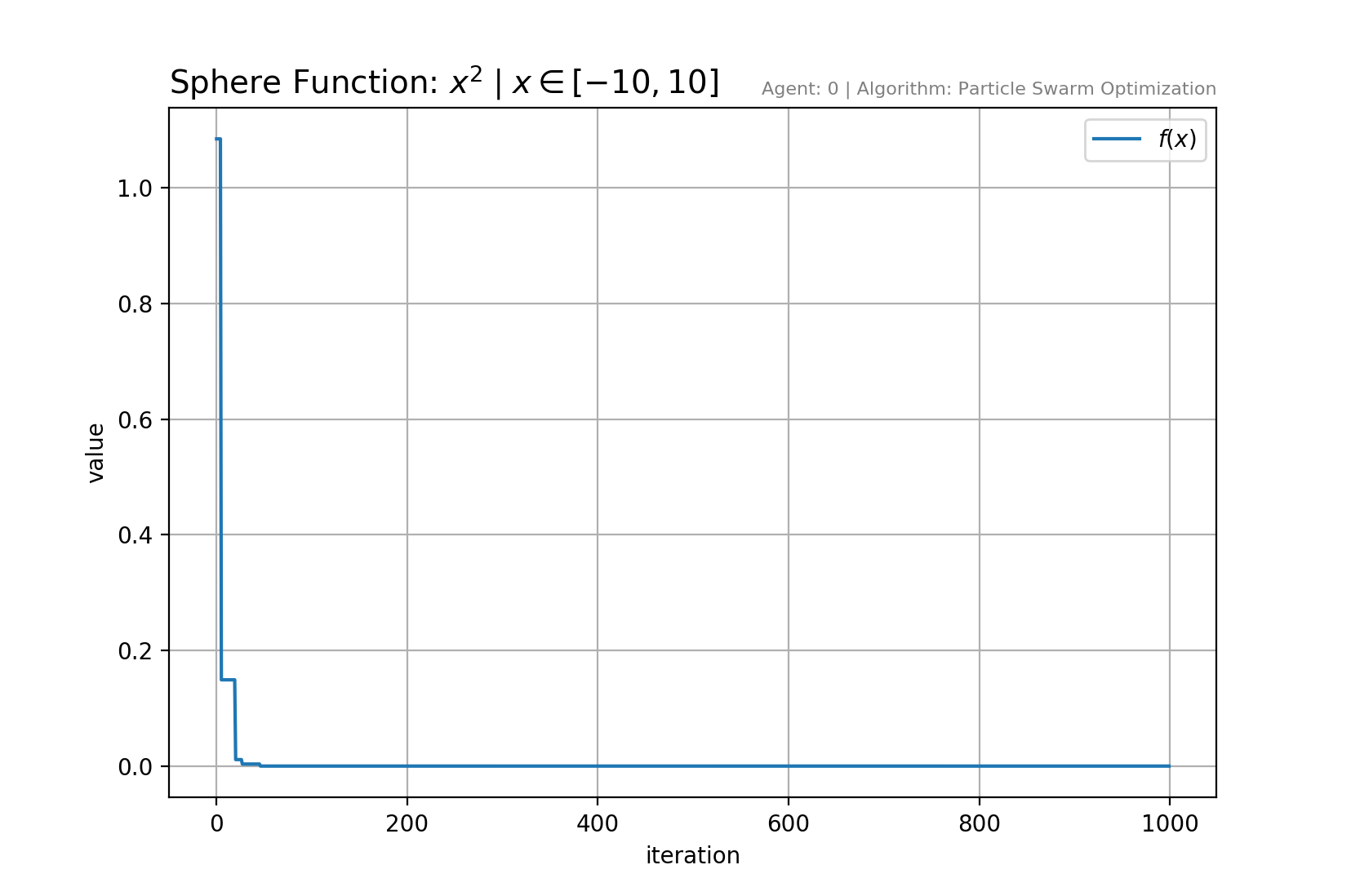}
\caption{Fitness' convergence plot using the visualization package.}
\label{f.convergence_fit}    
\end{figure}

Finally, the last package provides every visual-related method, where one can check a specific variable convergence, the fitness function convergence, and more. Currently, the visualization package implements the following module, which is also portrayed in Figure~\ref{f.flowchart_visualization}:

\begin{itemize}

\item \textbf{Convergence:} The convergence module provides a plotting function that helps users visualize their agents' positioning or fitness throughout the optimization task. Figures~\ref{f.convergence_pos} and~\ref{f.convergence_fit} illustrates some output plots that users can build.

\end{itemize}

\section{Library Usage}
\label{s.library}

In this section, we describe more deeply on how to work with Opytimizer. Fundamentally, one can read its documentation to learn more about the library or make use of the already-included examples. Moreover, we also provide methods to perform unitary tests and check if everything is working as expected.

\subsection{Installation}
\label{ss.installation}

We believe that everything has to be easy. Not tricky or daunting, Opytimizer will be the one-to-go package that you will need, from the very first installation to the daily-tasks implementing needs. If you may just run the following under your most preferred Python environment (raw, conda, virtualenv, whatever):

\verb|pip install opytimizer|

Alternatively, if you prefer to install the bleeding-edge version, please clone this repository and install it:

\verb|https://github.com/gugarosa/opytimizer.git|

\verb|pip install .|

Note that there is no additional requirement to use Opytimizer. As its unique dependencies are the Numpy and Matplotlib packages, it can be installed throughout every computer, regardless of the machine's operating system.

\subsection{Documentation}
\label{ss.documentation}

One might have an interest in learning the concepts and strategies behind Opytimizer more deeply. Therefore, we provide a fully documented reference\footnote{https://opytimizer.readthedocs.io} for everything that the library offers. From basic classes to more sophisticated methods, Opytimizer's documentation is the ideal source to understand how the library was built or even help us with further contributions.

\subsection{Classes and Methods Examples}
\label{ss.example}

Additionally, in the same \verb|examples/| folder, we provide examples for all packages that the library implements, such as:

\begin{itemize}

\item \textbf{Core:} \verb|create_agent.py|, \verb|create_function.py|, \verb|create_node.py|;
\item \textbf{Functions:} \verb|create_weighted_function.py|;
\item \textbf{Math:} \verb|benchmarking_functions.py|, \verb|calculate_hypercomplex_numbers.py|, \\ \verb|general_functions.py|, \verb|generate_distributions.py|, \verb|generate_random_numbers.py|;
\item \textbf{Optimizers:} \verb|create_abc.py|, \verb|create_aiwpso.py|, \verb|create_ba.py|, \verb|create_bha.py|, \verb|create_cs.py|, \verb|create_fa.py|, \verb|create_fpa.py|, \verb|create_gp.py|, \verb|create_hs.py|, \verb|create_ihs.py|, \verb|create_pso.py|, \verb|create_wca.py|;
\item \textbf{Spaces:} \verb|create_hyper_space.py|, \verb|create_search_space.py|, \verb|create_tree_space.py|;
\item \textbf{Utils:} \verb|load_history.py|, \verb|load_mnist.py|;
\item \textbf{Visualization:} \verb|convergence_plots.py|.

\end{itemize}

Each example is composed of high-level definitions of how to use predefined classes and methods. One can perceive that it provides a basic explanation on how to instantiate each class and which parameters should be used or not to accomplish their requirements.

\subsection{Test Suites}
\label{ss.test}

\begin{figure}[!ht]
\centering
\includegraphics[scale=1]{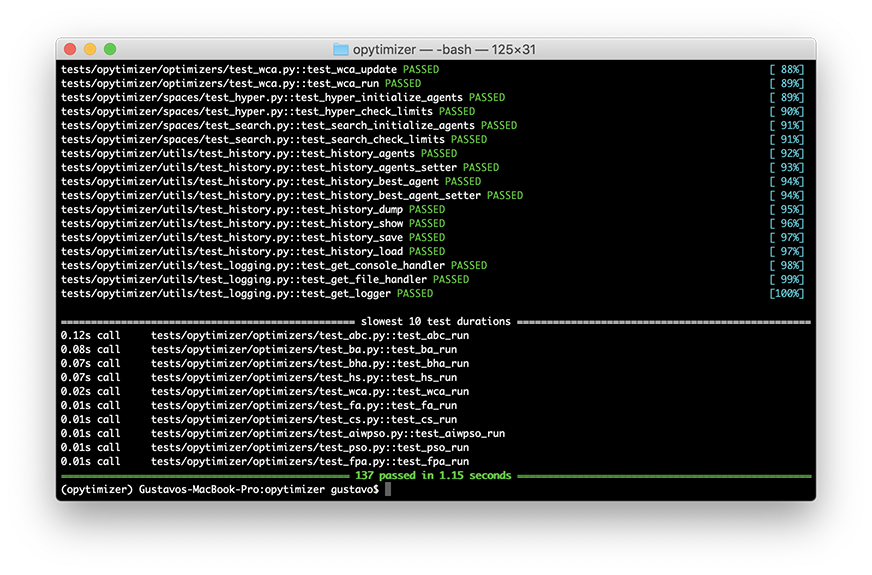}
\caption{Example of running tests with PyTest.}
\label{f.run_tests}    
\end{figure}

\begin{figure}[!ht]
\centering
\includegraphics[scale=1]{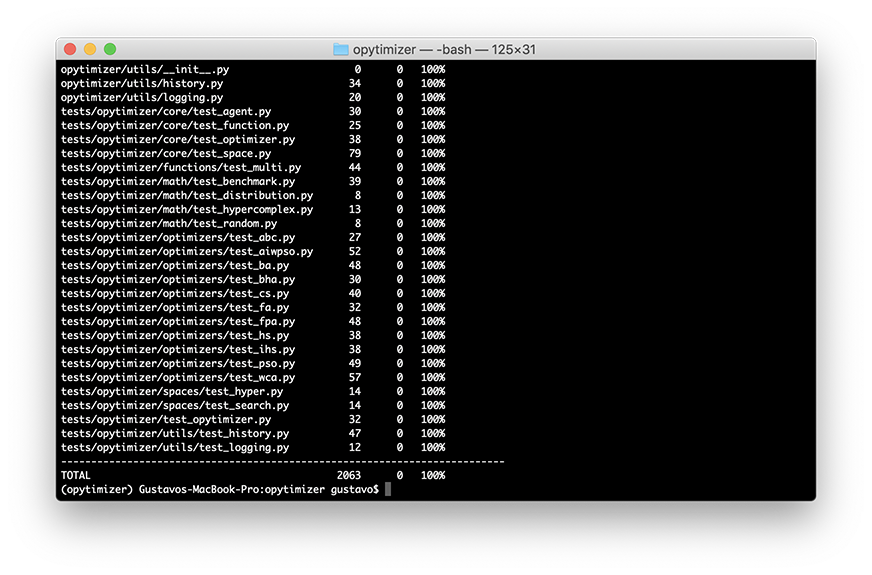}
\caption{Example of code coverage tests with Coverage.}
\label{f.coverage_tests}    
\end{figure}

Opytimizer is packed with tests to provide a more in-depth analysis of our code. Essentially, the idea behind any test is to check whether everything is working as expected or not. There are two main methods that we provide in order to run the tests:

\begin{itemize}
	\item \textbf{PyTest:} The first method that we offer is running solo the command \verb|pytest tests/|, as depicted by Figure~\ref{f.run_tests}. It will perform all the implemented tests and return an output declaring whether they passed or failed; 

	\item \textbf{Coverage:} An interesting addition to PyTest is the coverage module. Despite offering the same outputs from PyTest, it will also offer a report that documents how much the tests cover the code, as illustrated by Figure~\ref{f.coverage_tests}. Its usage is straightforward: \verb|coverage run -m pytest tests/| and \verb|coverage report -m|. 
\end{itemize}

\section{Applications}
\label{s.applications}

In this section, we describe within more details on how to run an application with Opytimizer. There are four pre-loaded applications included with the library, as well as, several integrations with well-known frameworks, such as PyTorch and Scikit-Learn.

\subsection{Getting Started}
\label{ss.getting_started}

After installing the Opytimizer library, it is straightforward to use its packages. Four basic examples show the main features which the library provides. One can refer to the \verb|examples/applications| folder and check the following files:

\begin{itemize}

\item \textbf{Single objective optimization:} \verb|single_objective_optimization.py|;
\item \textbf{Multi-objective optimization:} \verb|multi_objective_optimization.py|;
\item \textbf{Hyperspace optimization:} \verb|hyper_space_optimization.py|;
\item \textbf{Genetic Programming optimization:} \verb|genetic_programming_optimization.py|.

\end{itemize}

\begin{figure}[!ht]
\centering
\includegraphics[scale=1]{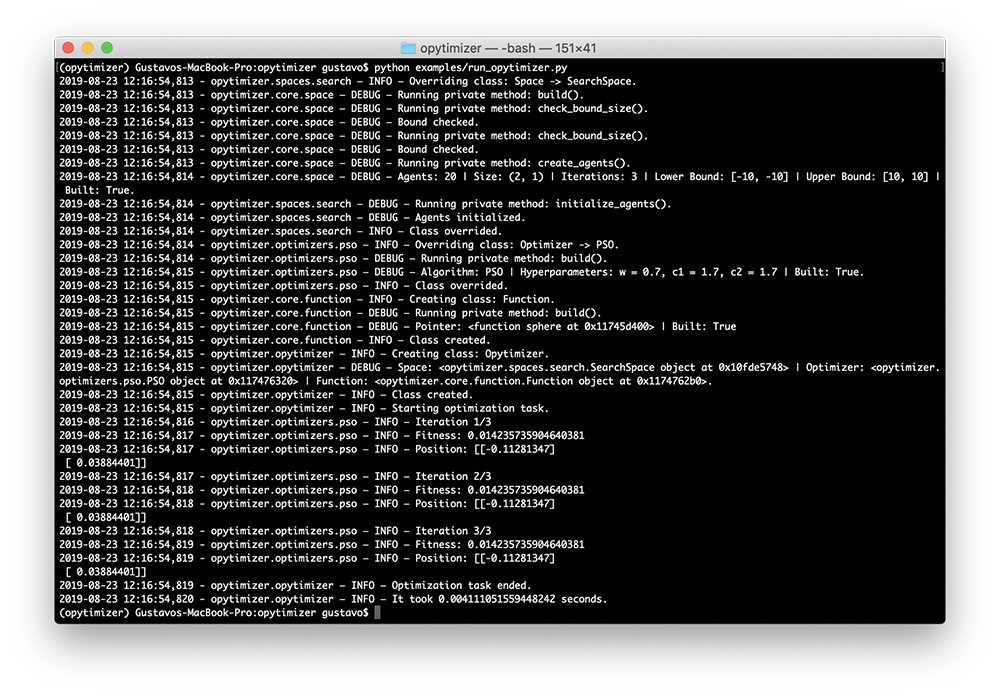}
\caption{Output logs generated by running an Opytimizer task.}
\label{f.get_started}    
\end{figure}

Each example is composed of a minimum set of variables to compose the search space, a mathematical function to be optimized\footnote{Note that the benchmark module provides some pre-defined functions.}, and an optimizer, which is the meta-heuristic technique used to perform the optimization process. Furthermore, they are bundled to an Opytimizer class, which holds all the vital information about the optimization task. Finally, the task is started, and when it finishes, it returns a history object which encodes valuable data about the optimization procedure. Figure~\ref{f.get_started} illustrates the output logs generated by an Opytimizer task.

The difference between single and multi-objective optimization consists of the number of functions to be optimized. While single optimization attempts to optimize one objective function, the multi optimization can employ more than one objective function. As for now, we only offer a weighted-sum approach to compose the multi-objective architecture, i.e., each objective function weighs towards the final objective function.

Additionally, the hyperspace optimization encodes agents in a hyperspace-based search space. Instead of using only a one-dimensional real-valued space, the hyperspace approach uses more dimensions throughout the optimization process. Commonly, there are three main hyperspaces that the literature proposes: (i) complex (2-dimensions), (ii) quaternions (4-dimensions), and (iii) octonions (8-dimensions).

Finally, Genetic Programming optimization is a tree-based evolutionary algorithm, which uses an entirely different structure than the previous examples. The idea is to provide a set of terminal and function (mathematical operators) nodes in order to compose the trees. A possible solution is harvested by traversing the tree and combining its node into a mathematical expression, which is further evaluated according to an objective function. Additionally, for each generation, trees are reproduced, crossed over, and mutated in order to compose the new population.

\subsection{Integrations}
\label{ss.integration}

Instead of modeling a new experiment, one can use our already predefined integrations to gather more knowledge on how to use Opytimizer or even on how to play with parameters. Inside the \verb|examples/integrations| folder, there are several well-known machine learning libraries used by the community. Each sub-folder contains diverse examples on how to integrate Opytimizer with a specific technique, e.g., how to optimize Convolutional Neural Networks parameters with a Particle Swarm Optimization. Currently, we offer the following integrations:

\begin{itemize}
	\item \textbf{LibOPF}\footnote{https://github.com/jppbsi/LibOPF}: \verb|opf_clustering.py|, \verb|opf_wrapper.py|;
	\item \textbf{PyTorch}\footnote{https://pytorch.org}: \verb|cnn.py|, \verb|enhanced_neural_network.py|, \verb|linear_regression.py|, \\ \verb|logistic_regression.py|, \verb|lstm.py|, \verb|neural_network.py|;
	\item \textbf{Recogners}\footnote{https://github.com/recogna-lab/recogners}: \verb|dropout_rbm.py|, \verb|rbm.py|;
	\item \textbf{Scikit-Learn}\footnote{https://scikit-learn.org}: \verb|k_means_clustering.py|, \verb|svm.py|.	
\end{itemize}

\begin{figure}[!ht]
\centering
\includegraphics[scale=0.85]{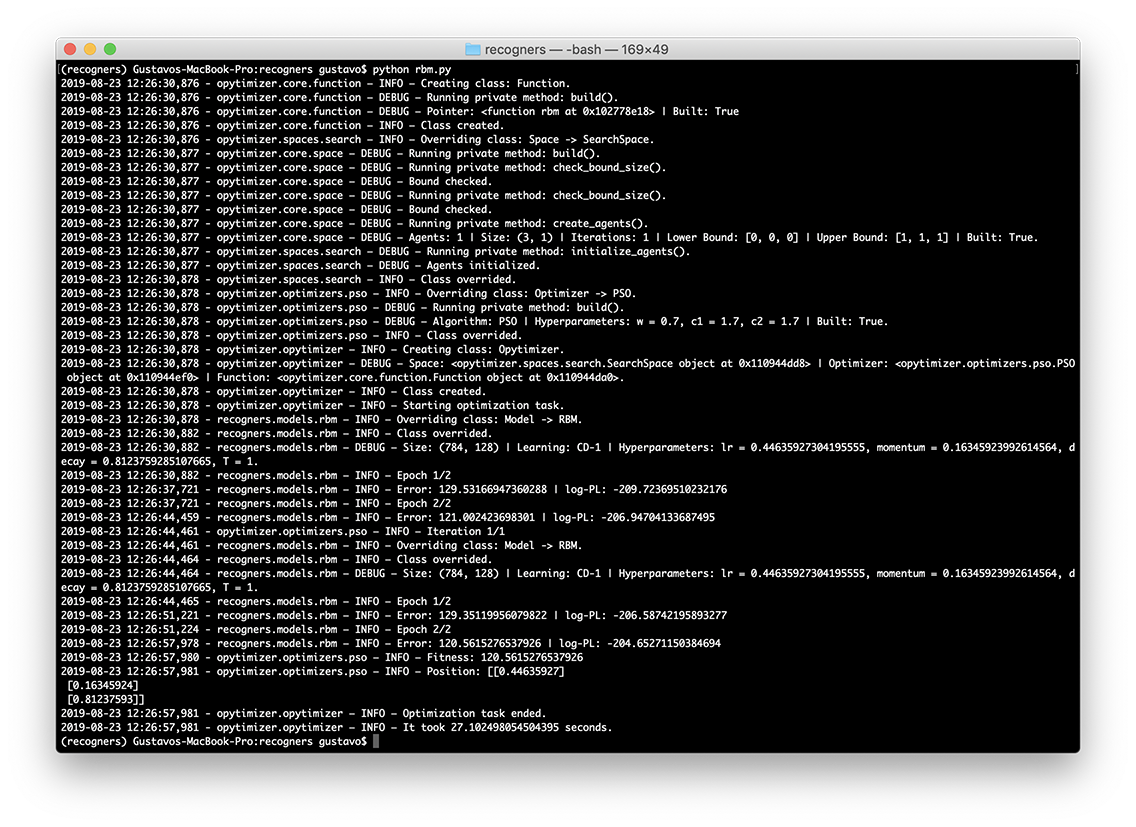}
\caption{Recogners' RBM optimization using Opytimizer.}
\label{f.rbm}    
\end{figure}

Each integration example is composed of two parts: (i) the objective function to be optimized, e.g., a convolutional neural network, a logistic regression, among others; and (ii) the minimum classes and methods to initialize the optimization procedure. For every example, the idea behind Opytimizer's functioning is almost the same: define the objective function, create a search space, instantiate an optimizer, and bundle everything into an Opytimizer class. Essentially, the main difference between all integrations scripts is the objective function to be optimized. Figure~\ref{f.rbm} illustrates an example of optimizing three parameters of a Restricted Boltzmann Machine (RBM): learning rate, momentum, and weight decay.

\subsection{Modeling a New Experiment}
\label{ss.function}

In order to model a new experiment, there are some standard rules that need to be followed. First of all, the objective function needs to be defined, which in this case we will be optimizing the well-known Sphere benchmarking function:

\begin{lstlisting}[language=Python]
import numpy as np
from opytimizer.core.function import Function

def sphere(x):
	# f(x) = x^2
	y = x ** 2
	
	return np.sum(y)
	
# Creating Function's object
f = Function(pointer=sphere)
\end{lstlisting}

Furthermore, we need to define the basic variables to initialize the search space:

\begin{lstlisting}[language=Python]
from opytimizer.spaces.search import SearchSpace

# Number of agents
n_agents = 20

# Number of decision variables
n_variables = 2

# Number of running iterations
n_iterations = 1000

# Lower and upper bounds (has to be the same size as n_variables)
lower_bound = [-5.12, -5.12]
upper_bound = [5.12, 5.12]

# Creating the SearchSpace class
s = SearchSpace(n_agents=n_agents, n_iterations=n_iterations,
				 n_variables=n_variables, lower_bound=lower_bound, 						upper_bound=upper_bound)	
\end{lstlisting}

Afterward, we need to define the hyperparameters of the optimizer and its class:

\begin{lstlisting}[language=Python]
from opytimizer.optimizers.pso import PSO

# Hyperparameters for the optimizer
hyperparams = {
    'w': 0.7,
    'c1': 1.7,
    'c2': 1.7
}

# Creating PSO's optimizer
p = PSO(hyperparams=hyperparams)	
\end{lstlisting}

Finally, we can bundle everything in the Opytimizer class and initialize the optimization task:

\begin{lstlisting}[language=Python]
from opytimizer import Opytimizer

# Finally, we can create an Opytimizer class
o = Opytimizer(space=s, optimizer=p, function=f)

# Running the optimization task
history = o.start()	
\end{lstlisting}

At the end of the optimization, a history object is produced, containing vital information about the task:

\begin{lstlisting}[language=Python]
# Displaying optimization's history
history.show()	
\end{lstlisting}

\section{Conclusions}
\label{s.conclusion}

In this article, we presented a nature-inspired open-source library for handling function optimization called Opytimizer. Using an object-oriented implementation, Opytimizer provides a sophisticated yet straightforward implementation, allowing users to prototype new meta-heuristic algorithms rapidly and integrating with other well-known libraries, such as LibOPF, PyTorch, Recogners, Scikit-Learn.

The library implements a broad range of meta-heuristic techniques, from evolutionary- to swarm-based methods, as well as a couple of benchmarking functions, used to validate the optimizations' performance. Additionally, as Opytimizer uses a tensor-based structure, it is possible to extend the idea of real-value optimization to more complex search spaces, the so-called hypercomplex optimization. Furthermore, Opytimizer provides a multi-objective optimization environment, empowering users with the ability to enhance their tasks and needs. Finally, Opytimizer also provides a history-based object, which used to track the whole optimization process and can be employed to furnish charts and insightful information about the procedure.

Regarding future works, we intend to make available more meta-heuristic algorithms and support more multi-objective techniques. Additionally, we also intend to create a more robust visualization package, allowing users to feed their history objects and furnish more charts. Moreover, we aim to add more benchmarking functions and integrations continually, providing a one-to-go optimization library for all purposes.

\section*{Acknowledgments}
The authors are grateful to FAPESP grants \#2013/07375-0, \#2014/12236-1, \#2017/25908-6, \#2018/15597-6, and \#2019/02205-5, as well as CNPq grants \#307066/2017-7 and \#427968/2018-6.

\bibliographystyle{unsrt}  
\bibliography{paper}

\end{document}